\documentclass{article}
\usepackage{ijcai22}

\usepackage{times}
\usepackage{soul}
\usepackage{url}
\usepackage[hidelinks]{hyperref}
\usepackage[utf8]{inputenc}
\usepackage[small]{caption}
\usepackage{graphicx}
\usepackage{amsmath}
\usepackage{amsthm}
\usepackage{booktabs}
\usepackage{algorithm}
\usepackage{algorithmic}
\usepackage{multirow}
\title{Community Question Answering Entity Linking via Leveraging Auxiliary Data}

\author{
Yuhan Li
\and
Wei Shen\footnote{Corresponding author}\and
Jianbo Gao\And
Yadong Wang
\affiliations
TMCC, TKLNDST, College of Computer Science, Nankai University, Tianjin, China\\
\emails
\{yuhanli, jianbo.gao, yadongwang\}@mail.nankai.edu.cn,
shenwei@nankai.edu.cn
}

\begin{document}

\maketitle

\begin{abstract}
  Community Question Answering (CQA) platforms contain plenty of CQA texts (i.e., questions and answers corresponding to the question) where named entities appear ubiquitously. In this paper, we define a new task of CQA entity linking (CQAEL) as linking the textual entity mentions detected from CQA texts with their corresponding entities in a knowledge base. This task can facilitate many downstream applications including expert finding and knowledge base enrichment. Traditional entity linking methods mainly focus on linking entities in news documents, and are suboptimal over this new task of CQAEL since they cannot effectively leverage various informative auxiliary data involved in the CQA platform to aid entity linking, such as parallel answers and two types of meta-data (i.e., topic tags and users). To remedy this crucial issue, we propose a novel transformer-based framework to effectively harness the knowledge delivered by different kinds of auxiliary data to promote the linking performance. We validate the superiority of our framework through extensive experiments over a newly released CQAEL data set against state-of-the-art entity linking methods.
\end{abstract}

% \vspace{-2mm}
\section{Introduction}

Community Question Answering (CQA) platforms, such as Quora\footnote{\url{https://www.quora.com}}, Yahoo! Answers\footnote{\url{https://answers.yahoo.com}}, and Stack Overflow\footnote{\url{https://www.stackoverflow.com}}, are Internet-based crowdsourcing services which enable users to post questions and seek answers from other users. The rapid growth of CQA platforms has attracted much research attention in recent years, such as answer ranking \cite{lyu2019we}, expert finding \cite{li2021askme}, and question retrieval \cite{ruckle2019improved}. The content in these platforms is usually organized as a question and a list of answers, associated with meta-data like topic categories of the question and users' information. In this paper, we define a question and all answers corresponding to this question as a CQA text. CQA texts are often ambiguous, especially with respect to the frequent occurrences of named entities. Specifically, a textual name in CQA texts may refer to many different entities in the real world, and a named entity may be expressed as various surface forms in CQA texts. Nevertheless, entity linking (EL), a popular text disambiguation task that aims to map entity mentions in text to their corresponding entities in a knowledge base, is rarely investigated in the CQA environment.

\begin{figure*}[t]
  \centering
  \includegraphics[width=1\textwidth]{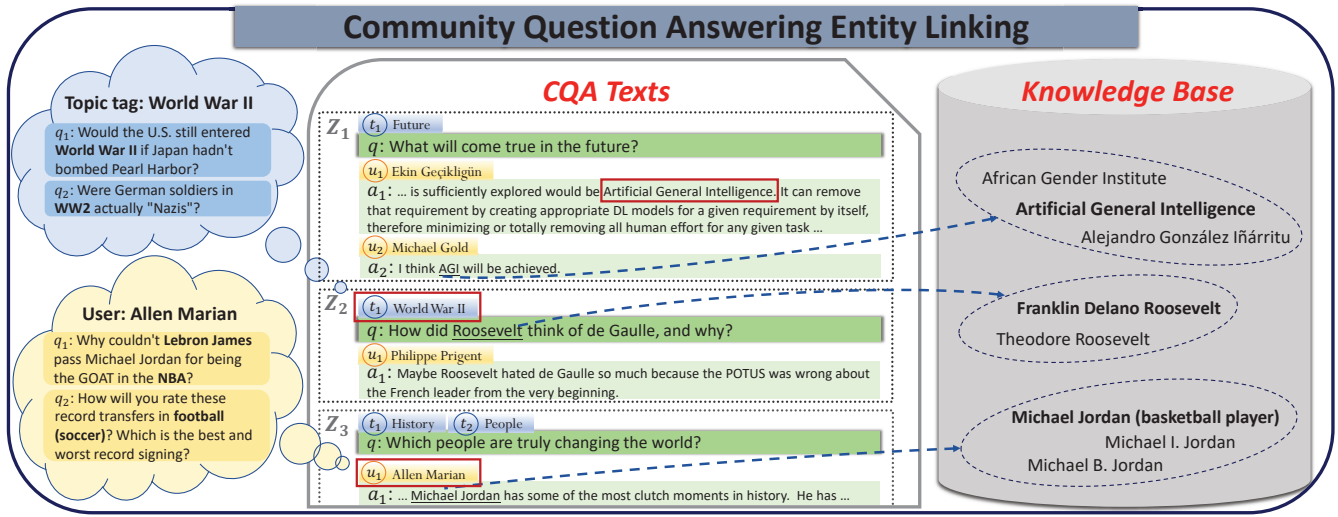}
  \caption{An illustration for the task of CQAEL. A CQA text is composed of a question and several parallel answers; A CQA text is accompanied with several topic tags, each of which may be involved in other questions; Each answer is associated with a user, who could ask or answer other questions; Entity mentions detected in CQA texts that need to be linked are \underline{underlined}; Candidate mapping entities in a knowledge base for each entity mention are shown via a dashed arrow line and circle; Correct corresponding entities are in \textbf{boldface}.}
  \label{fig:CQAEL}
  % \vspace{-4mm}
\end{figure*}

To derive a better understanding of CQA texts, we define a new task of Community Question Answering entity linking (CQAEL), as linking textual entity mentions detected from CQA texts with their corresponding named entities in a knowledge base (KB). For the example shown in Figure \ref{fig:CQAEL}, in the CQA text $Z_2$, the entity mention ``Roosevelt'' in question $q$ may refer to the $32$-th president of the United States ``Franklin Delano Roosevelt'', the $26$-th president of the United States ``Theodore Roosevelt'', or many other named entities which could be referred to as ``Roosevelt''. The goal of the CQAEL task is to link this ambiguous entity mention with its corresponding named entity ``Franklin Delano Roosevelt''.

CQAEL bridges CQA with KB, and also contributes to both. On one hand, CQAEL plays a fundamental role in a wide range of CQA tasks, such as expert finding \cite{li2021askme}, answer ranking \cite{lyu2019we}, and answer summarization \cite{song2017summarizing}. For example, finding users with expertise to answer some question is a central issue in the CQA platform. Linking ambiguous mentions in the question to their corresponding entities in KBs can further enhance the understanding of the question by leveraging the background knowledge of mapping entities provided by KBs. In such case, we can recommend the question to the best matching expert user. On the other hand, KB enrichment \cite{cao2020open} also benefits from CQAEL. To be specific, CQA texts embody invaluable knowledge about named entities. Developing an EL model capable of bridging CQA texts with KBs can help to enrich and update existing KBs significantly.

Nevertheless, CQA texts pose several challenges to entity linking. First, CQA texts (especially questions) are often concise and short, which makes them difficult to provide ample contextual information for context similarity computing in EL. Second, CQA texts are usually informal. Fortunately, however, CQA platforms involve various informative auxiliary data including parallel answers and two types of meta-data (i.e., topic tags and users), which have been proven effective for many CQA tasks \cite{tomasoni2010metadata,zhou2015learning}. We believe these different kinds of auxiliary data could supplement valuable knowledge beneficial for entity linking.

Specifically, sometimes an answer may be too short to offer sufficient context for disambiguating the entity mention in it. In such case, other answers under the same question (i.e., parallel answers) can be employed to enrich the context for the entity mention, as answers under the same question are usually relevant. We take the CQA text $Z_1$ in Figure \ref{fig:CQAEL} as an example. Answer $a_2$ in $Z_1$ is too short to supply enough context for linking mention ``AGI" correctly. Meanwhile, its parallel answer $a_1$ in $Z_1$ is much longer and contains the phrase ``Artificial General Intelligence'', i.e., the full name of the entity mention ``AGI'', which prompts the linking system to map ``AGI" with its correct entity via taking this parallel answer into account.
What's more, in the CQA platform, topic tags could be added to summarize basic topics in some keywords for each CQA text \cite{zhou2015learning}. In Figure \ref{fig:CQAEL}, the CQA text $Z_2$ is associated with topic tag $t_1$ ``World War II". As we know the fact that Franklin Delano Roosevelt is the president of the United States during the Second World War, the entity mention ``Roosevelt'' in question $q$ of $Z_2$ probably refers to the entity ``Franklin Delano Roosevelt'' rather than ``Theodore Roosevelt'' with the indication of this topic tag. 
Additionally, users usually ask or answer questions according to their individual interests or experience \cite{lyu2019we}. According to the questions that user $u_{1}$ named ``Allen Marian'' in the CQA text $Z_3$ has asked or answered in the past, we could know that user $u_{1}$ in $Z_3$ is interested in ``Sports". The entity mention ``Michael Jordan'' in answer $a_{1}$ of $Z_3$ that user $u_{1}$ has answered is likely to refer to the basketball player rather than the scientist ``Michael I. Jordan'' and the actor ``Michael B. Jordan'' with the indication of the user interest.

However, it is impractical and unreasonable to directly take the auxiliary data as an expansion of the context for the entity mention since the auxiliary data are numerous and noisy. For instance, there are usually more than thousands of questions under one topic tag. Moreover, a user is often interested in multiple domains, so questions they answered or asked are not all relevant to the entity mention that needs to be linked. Accordingly, the biggest challenge of the CQAEL task is how to effectively exploit the auxiliary data to aid entity linking. Traditional entity linking methods \cite{shen2021entity} primarily focus on linking entities in news documents, and are suboptimal over this new task since they do not consider how to make use of various informative auxiliary data specially existing in the CQA platform.

To deal with the above issue, we propose a novel transformer-based framework to harness the abundant knowledge embedded in different kinds of auxiliary data to improve entity linking. Specifically, a base module is derived to cross-encode the context of the entity mention and the candidate entity description to perform deep cross-attention between each other. To effectively exploit the auxiliary data, we propose an auxiliary data module, which is able to capture semantic relationships between the candidate entity description and different kinds of auxiliary data (i.e., parallel answers, topic tags, and users) in a unified manner. This auxiliary data module can provide effective and complementary linking evidence mined from auxiliary data and is flexible to be integrated with other EL models, which has been verified by our experiments. The main contributions of this paper are as follows.

\begin{itemize}
  \item We are among the first to explore the task of Community Question Answering entity linking (CQAEL), a new and important problem due to its broad applications.
  \item We propose a novel transformer-based framework which can leverage different kinds of auxiliary data in the CQA platform effectively to enhance the linking performance.
  \item We construct a finely-labeled data set named QuoraEL via crawling from Quora for the CQAEL task. Extensive experiments validate the superiority of our proposed framework against state-of-the-art EL methods. We release the data set and codes to facilitate the research towards this new task\footnote{\url{https://github.com/yhLeeee/CQA_EntityLinking}}.
\end{itemize}

% \vspace{-3mm}
\section{Task and Dataset}

% \vspace{-1mm}
\subsection{Task Definition}
\label{sec:taskdef}

A CQA text denoted by $Z$ includes a question $q$ and a set of its corresponding answers $A = \{a_1, a_2, ..., a_n\}$, where $a_i$ is the $i$-th answer of $q$ (1 $\le $ $i$ $\le $ $n$), i.e., $Z = \{q, A\}$. For each answer $a_i$, the other answers under the same question are considered as its parallel answers $\mathcal{A} = A - \{a_i\}$. Each CQA text is associated with a set of topic tags $T = \{t_1, t_2, ..., t_l\}$, where $t_j$ denotes the $j$-th topic tag (1 $\le j \le l$). We define topic meta-data as $MetaT = \{(t_j, Q_{t_j}) \mid  \forall t_j \in T\}$, where $Q_{t_j}$ is a set of questions involving topic tag $t_j$. Additionally, we denote the set of users as $U = \{u_1, u_2, ..., u_n\}$, where $u_i$ is the user who gives answer $a_i$. We define user meta-data as $MetaU = \{(u_i, Q_{u_i}) \mid \forall u_i \in U\}$, where $Q_{u_i}$ represents a set of questions asked or answered by user $u_i$. We regard $\mathcal{A}$, $MetaT$, and $MetaU$ as three kinds of auxiliary data involved in the CQA platform used in this paper.

Formally, given a CQA text $Z$ in which a set of entity mentions $M = \{m_1, m_2, ..., m_{|M|}\}$ are identified in advance and a KB containing a set of named entities $E = \{e_1, e_2, ..., e_{|E|}\}$, the task of CQAEL is to find a mapping $M \mapsto E$ that links each entity mention to its corresponding entity in the KB, via leveraging different kinds of auxiliary data in the CQA platform.

Before linking, for each entity mention, a small set of potential mapping entities are first chosen by candidate generation methods to prune the search space. Following the previous works \cite{phan2017neupl,fang2019joint}, we adopt the dictionary-based method to generate a set of candidate mapping entities $E_{m}$ for each entity mention $m$.

% \vspace{-1.5mm}
\subsection{Dataset Construction}

We create a new data set named QuoraEL to support the study of the CQAEL task. We choose Quora as the data source, which is one of the largest CQA platforms. We use a two-step data set collection process. For the first step, we extract CQA texts as well as their associated topic tags and users via crawling question pages of Quora. To collect two types of meta-data for CQA texts, in the second step we crawl topic meta-data (i.e., a set of questions involving each topic tag) and user meta-data (i.e., a set of questions asked or answered by each user) from topic pages and user homepages of Quora, respectively.

% Using the same Wikipedia dump (May 2019) as the one which \cite{wu2020scalable} used to get entity descriptions.
We use the July 2019 version of the Wikipedia dump as the reference KB. To annotate entity mentions appearing in the CQA text with their corresponding entities, we perform a two-stage labeling process with automatic labeling first and human labeling later. To be specific, we use the Stanford CoreNLP package\footnote{\url{https://stanfordnlp.github.io/CoreNLP/}} to automatically recognize and link entity mentions to their corresponding Wikipedia entities. Human annotators are introduced next to correct any false labels.

The final QuoraEL data set consists of $504$ CQA texts in total, in which there are $2192$ answers, $8030$ labeled entity mentions, and $1165$ topic tags. In average, each CQA text contains $4.35$ answers, $15.93$ labeled entity mentions, and $2.31$ topic tags. For the meta-data, we keep at most $10$ questions for each topic tag and $20$ questions for each user ($10$ questions asked and $10$ questions answered by the user).

% \begin{table}[t]
%   \centering
%     \begin{tabular}{lc}
%     \toprule
%     \# Total CQA texts & 504 \\
%     \# Total entity mentions & 8030 \\
%     \# Total answers & 2192 \\
%     \# Total topic tags & 1165 \\
%     \midrule
%     \# Average entity mentions per CQA text & 15.93 \\
%     \# Average answers per CQA text & 4.35 \\
%     \# Average topic tags per CQA text & 2.31 \\
%     \midrule
%     \# Max questions per topic tag & 10 \\
%     \# Max questions per user & 20 \\
%     \bottomrule
%     \end{tabular}%
%   \caption{Statistics of the created data set QuoraEL.}
%   \label{tab:QuoraEL}%
% \end{table}%

% To ensure the correctness of annotations, we manually recognize and annotate entity mentions with their corresponding entities.

% \vspace{-3mm}
\section{The Proposed Framework}
% \vspace{-1mm}

\begin{figure}[t]
  \centering
  \includegraphics[width=0.48\textwidth]{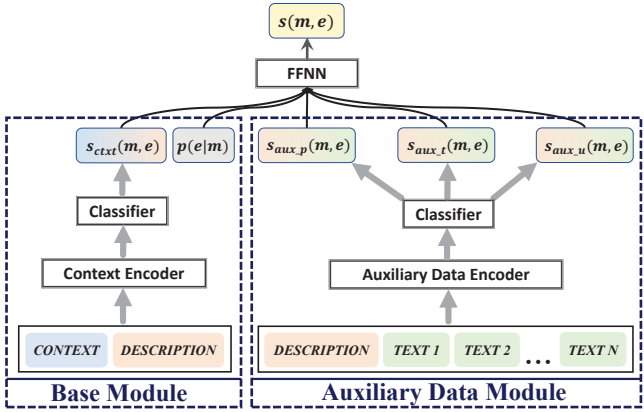}
  \caption{The overall architecture of our framework.}
  \label{fig:framework}
  % \vspace{-4mm}
\end{figure}

Our proposed framework is composed of two modules: a base module that mainly utilizes the context of the entity mention for disambiguation, and an auxiliary data module which employs different kinds of auxiliary data to facilitate linking in a unified manner. The overall architecture of our framework is shown in Figure \ref{fig:framework}.

% \vspace{-2mm}
\subsection{Base Module}

In the base module, two widely used entity linking features are considered, i.e., context similarity and prior probability.

\paragraph{Context Similarity.}
Measuring the similarity between the context of an entity mention (i.e., mention context) and the description text in an entity's Wikipedia page (i.e., entity description) is a crucial feature for entity linking. We propose to employ an XLNet-based cross-encoder to jointly encode the mention context and the candidate entity description to perform deep cross-attention between each other. XLNet \cite{yang2019xlnet}, a common transformer-based model which integrates the segment recurrence mechanism and relative encoding scheme of Transformer-XL \cite{dai2019transformer} into pretraining, has shown remarkable effect in many NLP tasks.

Formally, given an entity mention $m$, for each candidate entity $e \in E_{m}$, we first concatenate the mention context $\mathcal{C}$, the candidate entity description $\mathcal{D}$, and two special tokens from the vocabulary of XLNet as an input sequence. \texttt{[CLS]} indicates the start of the sequence, and \texttt{[SEP]} is added here to separate inputs from different segments. This concatenated string is encoded using XLNet to obtain $\mathbf{h}_{\mathcal{C}, \mathcal{D}}$, a representation for this context-description pair (from the \texttt{[CLS]} token). The context similarity feature $s_{ctxt}(m, e)$ is generated using a dot-product with a learned parameter vector $\mathbf{w}_1$ as follows. 

% \vspace{-3mm}
\begin{equation}
% \vspace{-1mm}
\begin{split}
  \textbf{h}_{\mathcal{C}, \mathcal{D}} = \textrm{XLNet}(\texttt{[CLS]} & \mathcal{C} \texttt{[SEP]} \mathcal{D} \texttt{[SEP]}), \\
   s_{ctxt}(m, e) &= \mathbf{w}_1^{T} \mathbf{h}_{\mathcal{C}, \mathcal{D}}
\end{split}  
\end{equation}

\paragraph{Combination with Prior Probability.}
Prior probability $p(e|m)$ is the probability of the appearance of a candidate entity $e$ given an entity mention $m$ without considering the context where the mention appears. It is estimated as the proportion of links with the mention form $m$ as anchor text pointing to the candidate entity $e$ in Wikipedia. We combine the context similarity feature with the prior probability feature as follows:

% \vspace{-3mm}
\begin{equation}
\label{eq:basemodel}
  s(m, e) = f(s_{ctxt}(m, e), p(e|m))
\end{equation}

\noindent where $f(\cdot)$ is a single layer feed-forward network and $s(m, e)$ is the ranking score for each candidate entity.

It is worth mentioning that the functionality of this base module is the same as most previous EL methods, all of which output a ranking score for each candidate entity based on the mention context. There are already massive effective EL solutions proposed in the past years \cite{shen2021entity}. Therefore, this base module is not the main focus of our paper and its design principle is simple and replaceable. Our experiments have validated that it is replaceable with other deep learning based EL models.

% \vspace{-2mm}
\subsection{Auxiliary Data Module}
\label{sec:auxiliarymodule}

Besides the mention context utilized by the aforementioned base module, there are various informative auxiliary data in the CQA platform, such as parallel answers and two types of meta-data (i.e., topic tags and users), which contain valuable knowledge beneficial for entity linking. Accordingly, we propose an auxiliary data module to effectively capture the knowledge delivered by different kinds of auxiliary data in a unified manner. This module can provide effective and complementary linking evidence mined from auxiliary data and be flexibly integrated with other EL models, which has been verified by our experiments.

In our task setting, each kind of auxiliary data is composed of multiple texts. For instance, parallel answers are a set of answer texts and both types of meta-data are sets of question texts. A na\"ive strategy is to take multiple texts as input directly and apply XLNet-based cross-encoder used in the base module once per text. However, auxiliary data are numerous. Directly applying this cross-encoder on multiple texts needs to encode each text-description pair separately which brings high time complexity. What's more, auxiliary data are noisy. Not all the texts in the auxiliary data are relevant to the given entity mention and helpful for its linking. Roughly treating them equally may incur noises.

To effectively exploit different kinds of auxiliary data, we start by selecting several useful texts from each auxiliary data to eliminate uninformative texts, and then another cross-encoder is introduced to jointly encode the candidate entity description and these selected useful texts in a single pass to derive the auxiliary data similarity.

\paragraph{Useful Text Selection.}
The basic idea is to select a subset of texts which are similar to the mention context from each kind of auxiliary data. The selected ones are regarded as useful texts which may expand the mention context and contribute to the linking process. Three common string similarity measures are adopted here, i.e, Difflib function, Jaro-Winkler similarity, and Levenshtein ratio. Given an entity mention with its associated mention context and a kind of auxiliary data, each text involved in the auxiliary data is scored via averaging these three string similarity measures between the mention context and the text. Ultimately, we regard the top-$k$ texts with the highest string similarity scores as useful texts for each kind of auxiliary data, where $k$ is a hyperparameter.

% \vspace{-1mm}
\paragraph{Auxiliary Data Similarity.}
Given the selected useful texts for each kind of auxiliary data, we regard them as a valid expansion of the mention context and calculate an auxiliary data similarity feature to indicate the proximity between this kind of auxiliary data and the candidate entity. Intuitively, the more the auxiliary data are semantically similar to the description text of a candidate entity, the more likely this candidate entity is the correct corresponding one. Unfortunately, the concatenation of the candidate entity description and these selected useful texts may usually exceed the token length limit of many transformer-based models such as XLNet. To tackle this issue, we propose to exploit a Longformer-based cross-encoder to jointly encode the candidate entity description and the selected useful texts to perform deep cross-attention between each other. Longformer \cite{beltagy2020longformer} is a modified transformer-based model that replaces the quadratic self-attention mechanism with a memory-efficient version, which combines local attention with sparse global attention. We choose Longformer due to the fact that it performs well in capturing semantic relationships between texts in one-many setting \cite{deyoung-etal-2021-ms} which is similar to ours, and it also allows us to encode thousands of tokens or more without high memory and time complexity.

In our setting, an entity mention $m$ is associated with three kinds of auxiliary data, each corresponding to a list of selected useful texts $q_1, q_2, ..., q_k$. Here, these useful texts are in descending order by their string similarity scores. As different kinds of auxiliary data have the same format (i.e., a list of texts), we process them in the same way. In the following, we take topic meta-data $MetaT$ as an example for illustration. For each candidate entity $e$, we concatenate the candidate entity description $\mathcal{D}$, selected useful texts of $MetaT$, and some special tokens as an input sequence. In Longformer, \texttt{[CLS]} and \texttt{[SEP]} are replaced by \texttt{<s>} and \texttt{</s>}. Additionally, \texttt{<d>}, \texttt{</d>} and \texttt{<q>}, \texttt{</q>} are special tokens representing description start and end, text start and end, respectively. The new special tokens are added to the model's vocabulary and randomly initialized before task fine-tuning. As suggested by \cite{beltagy2020longformer}, we assign global attention to the \texttt{<s>} token, and a sliding attention window of $64$ tokens allows each token to attend to its neighbors. We encode this sequence using Longformer to get $\mathbf{h}_{\mathcal{D}, MetaT}$, which is the output of the last hidden layer corresponding to \texttt{<s>}. The auxiliary data similarity feature $s_{aux\_t}(m, e)$ for topic meta-data is yielded via a dot-product with a learned parameter vector $\mathbf{w}_2$ as follows.

% \vspace{-4mm}
\begin{equation}
% \vspace{-2mm}
\begin{split}
  \textbf{h}_{\mathcal{D}, MetaT} = \textrm{L}&\textrm{ongformer}(\texttt{<s>}\texttt{<d>} \mathcal{D} \texttt{</d>} \texttt{</s>}\\
  &\texttt{<q>}q_1 \texttt{</q>} ... \texttt{<q>} q_k \texttt{</q>} \texttt{</s>}), \\
  s_{aux\_t}&(m, e) = \mathbf{w}_2^T \mathbf{h}_{\mathcal{D}, MetaT}
\end{split}
\end{equation}

\noindent Following the unified manner, we could derive the other two auxiliary data similarity features $s_{aux\_p}(m, e)$, $s_{aux\_u}(m, e)$ for parallel answers and user meta-data, respectively.

% \vspace{-2mm}
\subsection{Learning}

At present, we have obtained three kinds of auxiliary data similarity features. Thereafter, we extend Equation \ref{eq:basemodel} to Equation \ref{eq:auxmodel} via concatenating features involved in the base module with those three auxiliary data similarity features as follows:

% \vspace{-3mm}
\begin{equation}
\begin{split}
\label{eq:auxmodel}
  s(m, e) = f(&s_{ctxt}(m, e), p(e|m), \\
  &s_{aux\_p}(m, e), s_{aux\_t}(m, e), s_{aux\_u}(m, e))
\end{split}
\end{equation}

\noindent Subsequently, we normalize $s(m, e)$ using a softmax function and choose the candidate entity $e^*$ with the highest ranking score as the corresponding entity for the entity mention $m$ based on the following formulas:

% \vspace{-3mm}
\begin{equation}
  \hat{s}(m, e) =\frac{ \exp(s(m, e)) }{\sum_{e' \in E_m}\exp(s(m, e')) }
\end{equation}

\begin{equation}
  e^* = \mathop{argmax}\limits_{e' \in E_{m}} (\hat{s}(m, e'))
\end{equation}
% \vspace{-2mm}

\noindent where $E_m$ is a set of candidate entities for the entity mention $m$. We utilize a cross-entropy loss for training, which is defined as follows:

% \vspace{-2mm}
\begin{equation}
  \mathcal{L} = \sum_{Z \in \mathcal{Z}}^{} \sum_{m \in M}^{} \sum_{e \in E_m}^{} -(y\log \hat{s}(m, e))
\end{equation}
% \vspace{-2mm}

\noindent where $\mathcal{Z}$ denotes a training set of CQA texts and $y \in \{0, 1\}$ denotes the actual label of the candidate entity. If the candidate entity $e$ is the correct corresponding entity for the entity mention $m$, the value of $y$ is $1$; otherwise $0$. The goal of learning is to minimize the loss $\mathcal{L}$.

% \vspace{-2.5mm}
\section{Experiments}

\subsection{Experimental Setup}

We perform experiments on the newly created QuoraEL data set. We use 5-fold cross-validation and split the CQA texts into training ($70$\%), validation ($10$\%), and testing ($20$\%). For training, we adopt AdamW \cite{loshchilov2018decoupled} optimizer with a warmup rate $0.1$, an initial learning rate $1$e-$5$, and a mini-batch size $2$. Dropout with a probability of $0.1$ is used to alleviate over-fitting. For XLNet and Longformer, their parameters are initialized by the xlnet-base-cased and longformer-base-4096 models, respectively. For the base module, the maximum sequence length is set to $128$. We also experimented with $256$, which results in negligible improvement. For the auxiliary data module, maximum lengths of the candidate entity description and each text are set to $128$ and $64$, respectively. The hyperparameter $k$ is set to $3$, whose impact to the performance will be studied later. Based on our task definition, entity mentions have been recognized in advance and given as the input of the task. Therefore, we adopt accuracy as the evaluation metric, calculated as the number of correctly linked entity mentions divided by the total number of all the input entity mentions, the same as many state-of-the-art EL methods \cite{shen2021entity}. After training $10$ epochs, we select the model with the best performance on the validation set and evaluate its performance on the test set. All experiments are implemented by MindSpore Framework\footnote{\url{https://www.mindspore.cn/en}} with two NVIDIA Geforce GTX 3090 (24GB) GPUs.

\begin{table}[t]
  % \small
  \centering
    \begin{tabular}{l|c|c}
    \toprule
    \textbf{Models} & Base Setting & Aux Setting \\
    \midrule
    Deep-ED (2017) & 82.56 & 82.97 \\
    Ment-Norm (2018) & 82.99 & 83.19 \\
    Zeshel (2019) & 88.72 & 88.91 \\
    REL (2020) & 80.49 & 81.05 \\
    FGS2EE (2020) & 82.59 & 83.07 \\
    BLINK (2020) & 87.97 & 87.92 \\
    GENRE (2021) & 86.26 & 87.06 \\
    \midrule
    Base Module (ours) & \textbf{89.37} & - \\
    Full Module (ours) & - & \textbf{92.02} \\
    \bottomrule
    \end{tabular}%
  \caption{Effectiveness performance.}
  \label{tab:effectivenessstudy}%
  % \vspace{-4mm}
\end{table}%

% \vspace{-1.5mm}
\subsection{Effectiveness Study}

We compared our proposed framework with the following state-of-the-art EL models:

\begin{itemize}
\setlength{\itemsep}{0pt}
\item \textbf{Deep-ED} \cite{ganea2017deep} utilizes CRF for joint entity linking and solves the global training problem via truncated fitting LBP.
\item \textbf{Ment-Norm} \cite{le2018improving} improves Deep-ED by encoding relations between mentions via latent variables.
% \item \textbf{NCEL} \cite{cao2018neural} applies Graph Convolutional Network to perform global entity linking.
\item \textbf{Zeshel} \cite{logeswaran2019zero} leverages BERT cross-encoder to address EL in the zero-shot setting.
\item \textbf{REL} \cite{van2020rel} is one of the most prominent open source toolkits for EL.
\item \textbf{FGS2EE} \cite{hou2020improving} injects fine-grained semantic information into entity embeddings to facilitate EL.
\item \textbf{BLINK} \cite{wu2020scalable} utilizes both BERT bi-encoder and cross-encoder for entity linking.
\item \textbf{GENRE} \cite{de2021autoregressive} introduces a sequence-to-sequence EL model based on BART \cite{lewis2020bart}.
\end{itemize}

Table \ref{tab:effectivenessstudy} shows the linking accuracy of all models, in which full module denotes the combination of our base module and our auxiliary data module. The linking performance of all baselines is obtained via running their open-source solutions except REL, which provides a publicly available API. To give a thorough evaluation, we conduct comparison in two settings. One is called Base setting where only mention context is accessible and the other is called Aux setting where auxiliary data is accessible as well. Base setting is the same as traditional EL setting, and Aux setting is a special setting in the CQA environment. For all baselines in the Aux setting, to ensure fair comparison, we expand the mention context using the three kinds of auxiliary data and keep their model input consistent with our framework. 

In summary, our transformer-based framework consistently surpasses all the state-of-the-art EL baselines in both settings, demonstrating the superiority of our framework in tackling the CQAEL task. To be specific, our base module achieves better performance than several competitive baselines such as Zeshel and BLINK, which are elaborately designed for modeling context similarity, since these two baselines do not leverage prior probability. In addition, we could observe that all the baselines yield slightly better or even worse results after adding auxiliary data. This may be attributed to the fact that auxiliary data are noisy, and these traditional EL models cannot effectively capture the useful knowledge scattered in the auxiliary data to aid EL. Nevertheless, our full module promotes by $2.65$ absolute percentages via integrating our auxiliary data module, exhibiting its effectiveness in leveraging the auxiliary data to enhance the linking performance. 

\begin{table}[t]
  \centering
  \small
    \begin{tabular}{llcc}
    \toprule
    \multicolumn{2}{l}{\multirow{2}[4]{*}{\textbf{Models}}} & \multicolumn{2}{c}{Accuracy (\%)} \\
\cmidrule{3-4}    \multicolumn{2}{l}{} & Total & $\bigtriangleup$  \\
    \midrule
    \multicolumn{2}{l}{Base Module} & 89.37   & - \\
    \multicolumn{2}{l}{\quad $+$ \textit{Parallel answers}} & 91.55     & +2.18 \\
    \multicolumn{2}{l}{\quad $+$ \textit{User}} & 91.26     & +1.89 \\
    \multicolumn{2}{l}{\quad $+$ \textit{Topic}} & 91.77    & +2.40 \\
    \multicolumn{2}{l}{\quad $+$ \textit{User, Parallel answers}} & 91.61    & +2.24 \\
    \multicolumn{2}{l}{\quad $+$ \textit{Topic, User}} & 91.89     & +2.52 \\
    \multicolumn{2}{l}{\quad $+$ \textit{Topic, Parallel answers}} & 91.76     & +2.39 \\
    \midrule
    \multicolumn{2}{l}{Full Module} & 92.02     & +2.65 \\
    \midrule
    \midrule
    \multicolumn{2}{l}{Deep-ED \cite{ganea2017deep}} & 82.56   & - \\
    \multicolumn{2}{l}{\quad $+$ \textit{Our Auxiliary Data Module}} & 88.16     & +5.60 \\
    \midrule
    \midrule
    \multicolumn{2}{l}{Zeshel \cite{logeswaran2019zero}} & 88.72   & - \\
    \multicolumn{2}{l}{\quad $+$ \textit{Our Auxiliary Data Module}} &  91.49   & +2.77 \\
    \bottomrule
    \end{tabular}%
  % \vspace{-2mm}
  \caption{Ablation performance.}
  \label{tab:ablationstudy}%
  % \vspace{-5mm}
\end{table}%

% \vspace{-2mm}
\subsection{Ablation Study}

At the top of Table \ref{tab:ablationstudy}, we report the performance of our framework leveraging different kinds of auxiliary data. From the experimental results, we can see that each kind of auxiliary data has a positive contribution to the linking performance, and when all the three kinds of auxiliary data are consolidated together in our full module, it yields the best performance.

To study the effectiveness and flexibility of our auxiliary data module, we replace our base module with two popular traditional EL models (i.e., Deep-ED and Zeshel) and make them work with our auxiliary data module, whose results are shown at the bottom of Table \ref{tab:ablationstudy}. Specifically, the ranking score output by the EL model is concatenated with the three auxiliary data similarity features of our auxiliary data module, and then passed through a single layer feed-forward network, similar to Equation \ref{eq:auxmodel}. It can be seen from Table \ref{tab:ablationstudy} that we can obtain significant performance gain when our auxiliary data module is combined with each of the two traditional EL models, which clearly demonstrates the effectiveness and flexibility of our auxiliary data module. This confirms that our auxiliary data module indeed provides effective and complementary linking evidence delivered by different kinds of auxiliary data to help not only our base module, but also other EL models to promote their linking performance.

\begin{figure}[t]
  \centering
  \includegraphics[width=0.29\textwidth]{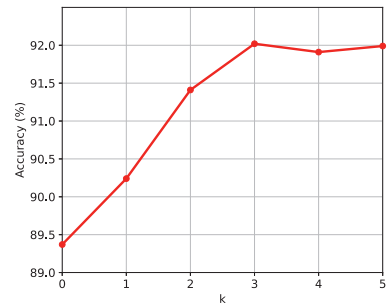}
  % \vspace{-2mm}
  \caption{Effects of the hyperparameter $k$.}
  \label{fig:hyper}
  % \vspace{-4mm}
\end{figure}

% \vspace{-2mm}
\subsection{Hyperparameter Study}

To investigate the effect of the hyperparameter $k$, i.e., the number of selected useful texts in our auxiliary data module, we conduct experiments with $k$ varied from $0$ to $5$, and plot the results in Figure \ref{fig:hyper}. We can see that our framework generally performs better with a larger $k$, and achieves high and stable accuracy (about $0.92$) when $k \ge 3$. This is due to the fact that more selected useful texts bring more helpful linking evidence, which may lead to better performance.

\section{Related Work}
% \vspace{-2mm}

Entity linking has gained increasing attention as a fundamental task to bridge unstructured text with knowledge bases, such as Wikipedia and Freebase. It acts as an important pre-processing step for many downstream knowledge-driven applications. A typical EL system often consists of two stages: candidate generation and entity ranking \cite{shen2014entity}. 
During the entity ranking stage, the key is to measure the similarity between the mention context and the entity description. Early EL works \cite{ratinov2011local} employ hand-crafted features to model this textual coherence. With the extensive application of deep learning, EL studies resort to neural network based methods. Bi-encoders based on CNN \cite{xue2019neural}, RNN \cite{fang2019joint}, and pre-trained language models \cite{fang2020high} are leveraged to encode those two text parts (i.e., mention context and entity description) individually. In this case, the semantic relationships between these two text parts are not fully exploited. Recent EL works \cite{wu2020scalable,tang2021bidirectional} employ BERT to cross-encode these two text parts to perform deep cross-attention between each other and achieve advanced performance.

Besides common news documents, entity mentions also appear in multi-source heterogeneous data. Recently, entity linking for web tables \cite{bhagavatula2015tabel}, tweets \cite{ran2018attention}, open knowledge bases \cite{liu2021joint}, and multimodal data \cite{gan2021multimodal} have been successively explored. \cite{wang2017named} only links entity mentions in the question of CQA and does not consider two types of meta-data (i.e., topic tags and users) specially existing in the CQA platform, which is different from ours. 

% \vspace{-2mm}
\section{Conclusion}
In this paper, we present a new task of CQAEL and create a data set QuoraEL to foster further study. We propose a novel transformer-based framework which can leverage different kinds of auxiliary data involved in the CQA platform effectively. An auxiliary data module is proposed to provide effective and complementary linking evidence mined from auxiliary data and is flexibly integrated with other EL models. Extensive experiments demonstrate the effectiveness of our framework against state-of-the-art EL methods.

% \vspace{-2mm}
\section*{Acknowledgments}

This work was supported in part by National Natural Science Foundation of China (No. U1936206), YESS by CAST (No. 2019QNRC001), and CAAI-Huawei MindSpore Open Fund.

{\small
\bibliographystyle{named}
\bibliography{ijcai22}
}
\end{document}